\documentclass{article}

\PassOptionsToPackage{numbers, compress}{natbib}

\usepackage[final]{neurips_2022}

\usepackage[utf8]{inputenc} 
\usepackage[T1]{fontenc}    
\usepackage{hyperref}       
\usepackage{url}            
\usepackage{booktabs}       
\usepackage{amsfonts}       
\usepackage{nicefrac}       
\usepackage{microtype}      
\usepackage{xcolor}         
\usepackage{caption}
\usepackage{graphicx}
\usepackage{bbm}

\usepackage[ruled,vlined]{algorithm2e}
\usepackage{algpseudocode}
\usepackage{multirow}
\usepackage{subfig}

\usepackage[tbtags]{amsmath}

\newcommand{\set}[1]{\mathcal{#1}}

\title{
Revisiting Realistic Test-Time Training: Sequential Inference and Adaptation by Anchored Clustering
}

\author{
  Yongyi Su$^{1}$ \quad Xun Xu$^{2}$
  \quad Kui Jia$^{13}$\\
  $^1$South China University of Technology \quad
  $^2$Institute for Infocomm Research \\
  $^3$Peng Cheng Laboratory \\
  \texttt{eesuyongyi@mail.scut.edu.cn} \\
  \texttt{alex.xun.xu@gmail.com} \\
  \texttt{kuijia@scut.edu.cn}
}

\begin{document}

\maketitle

\begin{abstract}
  Deploying models on target domain data subject to distribution shift requires adaptation. Test-time training~(TTT) emerges as a solution to this adaptation under a realistic scenario where access to full source domain data is not available and instant inference on target domain is required. Despite many efforts into TTT, there is a confusion over the experimental settings, thus leading to unfair comparisons. In this work, we first revisit TTT assumptions and categorize TTT protocols by two key factors. Among the multiple protocols, we adopt a realistic sequential test-time training~(sTTT) protocol, under which we further develop a \textit{test-time anchored clustering (TTAC)} approach to enable stronger test-time feature learning. TTAC discovers clusters in both source and target domain and match the target clusters to the source ones to improve generalization. Pseudo label filtering and iterative updating are developed to improve the effectiveness and efficiency of anchored clustering. We demonstrate that under all TTT protocols TTAC consistently outperforms the state-of-the-art methods on {six} TTT datasets. We hope this work will provide a fair benchmarking of TTT methods and future research should be compared within respective protocols. A demo code is available at \url{https://github.com/Gorilla-Lab-SCUT/TTAC}.
\end{abstract}

\section{Introduction}

The recent success in deep learning is attributed to the availability of large labeled data~\cite{krizhevsky2012imagenet,zhou2018brief} and the assumption of i.i.d. between training and test datasets. Such assumptions could be violated when test data features a drastic difference from the training data, e.g. training on synthetic images and test on real ones, and this is often referred to as domain shift~\cite{quinonero2008dataset,ben2010theory}. To tackle this issue, domain adaptation~(DA)~\cite{wang2018deep} emerges and the labeled training data and unlabeled testing data are often referred to as source and target data/domains respectively.

The existing DA works either require the access to both source and target domain data during training~\cite{ganin2015unsupervised} or training on multiple domains simultaneously~\cite{zhou2021domain}. The former approach renders the methods restrictive to limited scenarios where source domain data is always available during adaptation while the latter ones are computationally more expensive. 
To alleviate the reliance on source domain data, which may be inaccessible due to privacy issues or storage overhead, source-free domain adaptation (SFDA) emerges which handles DA on target data without access to source data~\cite{pmlr-v119-liang20a,kundu2020universal,yang2021generalized,xia2021adaptive,liu2021ttt++}. SFDA is often achieved through self-training~\cite{pmlr-v119-liang20a}, self-supervised learning~\cite{liu2021ttt++} or introducing prior knowledge~\cite{pmlr-v119-liang20a} and it requires multiple training epochs on the full target data to allow convergence. Despite easing the dependence on source data, SFDA has major drawbacks in a more realistic domain adaptation scenario where test data arrives in a stream and inference or prediction must be taken instantly, and this setting is often referred to as test-time training (TTT) or adaptation (TTA)~\cite{sun2020test,wang2020tent,iwasawa2021test,liu2021ttt++}. Despite the attractive feature of adaption at time test, we notice a confusion of what defines a test-time training and as a result comparing apples and oranges happens frequently in the community. In this work, we first categorize TTT by two key factors after summarizing various definitions made in existing works. First, under a realistic TTT setting, test samples are sequentially streamed and prediction must be made instantly upon the arrival of a new test sample. More specifically, the prediction of test sample $X_T$, arriving at time stamp $T$, should not be affected by any subsequent samples, $\{X_t\}_{t=T+1\cdots\infty}$. Throughout this work, we refer to the sequential streaming as \textbf{one-pass adaptation} protocol and any other protocols violating this assumption are called \textbf{multi-pass adaptation} (model may be updated on all test data for multiple epochs before inference). Second, we  notice some recent works must \textbf{modify source domain training loss}, e.g. by introducing additional self-supervised branch, to allow more effective TTT~\cite{sun2020test,liu2021ttt++}. This will introduce additional overhead in the deployment of TTT because re-training on some source dataset, e.g. ImageNet, is computationally expensive.   
In this work, we aim to tackle on the most realistic and challenging TTT protocol, i.e. one-pass test time training with no modifications to training objective. This setting is similar to TTA proposed in~\cite{wang2020tent} except for not restricting access to a light-weight information from the source domain. Given the objective of TTT being efficient  adaptation at test-time, this assumption is computationally efficient and improves TTT performance substantially. We name this new TTT protocol as \textbf{sequential test time training (sTTT)}. 

We propose two techniques to enable efficient and accurate sTTT. i) We are inspired by the recent progresses in unsupervised domain adaptation~\cite{tang2020unsupervised} that encourages testing samples to form clusters in the feature space. However, separately learning to cluster in the target domain without regularization from source domain does not guarantee improved adaptation~\cite{tang2020unsupervised}. 
To overcome this challenge, we identify clusters in both the source and target domains through a mixture of Gaussians with each component Gaussian corresponding to one category. Provided with the category-wise statistics from source domain as anchors, we match the target domain clusters to the anchors by minimizing the KL-Divergence as the training objective for sTTT. Therefore, we name the proposed method  \textit{test-time anchored clustering (TTAC)}. Since test samples are sequentially streamed, we develop an exponential moving averaging strategy to update the target domain cluster statistics to allow gradient-based optimization. ii) Each component Gaussian in the target domain is updated by the test sample features that are assigned to the corresponding category. Thus, incorrect assignments (pseudo labels) will harm the estimation of component Gaussian. To tackle this issue, we are inspired by the correlation between network's stability and confidence and pseudo label accuracy~\cite{lee2013pseudo,sohn2020fixmatch}, and propose to filter out potentially incorrect pseudo labels. Component Gaussians are then updated by the samples that have passed the filtering. To exploit the filtered out samples, we  incorporate a global feature alignment~\cite{liu2021ttt++} objective. 
We also demonstrate TTAC is compatible with existing TTT techniques, e.g. contrastive learning branch~\cite{liu2021ttt++}, if source training loss is allowed to be modified. The contributions of this work are summarized as below.

\begin{itemize}
\item In light of the confusions within TTT works, we provide a categorization of TTT protocols by two key factors. Comparison of TTT methods is now fair within each category.
\item We adopt a realistic TTT setting, namely sTTT. To improve test-time feature learning, we propose TTAC by matching the statistics of the target clusters to the source ones. The target statistics are updated through moving averaging with filtered pseudo labels. 
\item The proposed method is complementary to existing TTT method and is demonstrated on {six} TTT datasets, achieving the state-of-the-art performance under all categories of TTT protocols.
\end{itemize}
\vspace{-0.3cm}

\section{Related Work}

\noindent\textbf{Unsupervised Domain Adaptation}. 
Domain adaptation aims to improve model generalization when source and target data are not drawn i.i.d. When target data are unlabeled, unsupervised domain adaptation~(UDA)~\cite{ganin2015unsupervised,tzeng2014deep} learns domain invariant feature representations on both source and target domains to improve generalization. Follow-up works improve UDA by minimizing a divergence~\cite{gretton2012kernel,sun2016deep,zellinger2017central}, adversarial training~\cite{hoffman2018cycada} or discovering cluster structures in the target data~\cite{tang2020unsupervised}. Apart from formulating UDA as a task-specific model, re-weighting has been adopted for domain adaptation by selectively up-weighting conducive samples in the source domain~\cite{jiang2007instance, yan2017mind}. During model training, the existing approaches often require access to the source domain data which, however, may be not accessible due to privacy issues, storage overhead, etc. Therefore, deploying UDA in more realistic scenarios has inspired research into source-free domain adaptation and test-time training/adaptation.

\noindent\textbf{Source-Free Domain Adaptation}. 
Without the access to source data, source-free domain adaptation (SFDA) develops domain adaptation through self-training~\cite{pmlr-v119-liang20a, kundu2020universal, iwasawa2021test}, self-supervised training~\cite{liu2021ttt++}, clustering in the target domain~\cite{yang2021generalized} and feature restoration~\cite{eastwood2022sourcefree}. It has been demonstrated that SFDA performs well on seminal domain adaptation datasets even compared against UDA methods~\cite{tang2020unsupervised}. Nevertheless, SFDA requires access to all testing data beforehand and model training must be carried out iteratively on the testing data. In a more realistic DA scenario where inference and adaptation must be implemented simultaneously, SFDA will no longer be effective. Moreover, some statistical information on the source domain does not pose privacy issues and can be exploited to further improve adaptation on target data.

\noindent\textbf{Test-Time Training}. 
Collecting enough samples from target domain and adapt models in an offline manner restricts the application to adapting to a static target domain. To allow fast and online adaptation, test-time training (TTT)~\cite{sun2020test,wang2022continual} or adaptation (TTA)~\cite{wang2020tent} emerges. Despite many recent works claiming to be test-time training, we notice a severe confusion over the definition of TTT. In particular, whether training objective must be modified~\cite{sun2020test,liu2021ttt++} and whether sequential inference on target domain data is possible~\cite{wang2020tent,iwasawa2021test}. Therefore, to reflect the key challenges in TTT, we define a setting called sequential test-time training (sTTT) which neither modifies the training objective nor violates  sequential inference. Under the more clear definition, some existing works, e.g. TTT~\cite{sun2020test} and TTT++~\cite{liu2021ttt++} is more likely to be categorized into SFDA. Several existing works~\cite{wang2020tent,iwasawa2021test} can be adapted to the sTTT protocol. Tent~\cite{wang2020tent} proposed to adjust affine parameters in the batchnorm layers to adapt to target domain data. Nevertheless, updating only a fraction of model weights inevitably leads to limited performance gain on the target domain. T3A~\cite{iwasawa2021test} further proposed to update classifier prototype through pseudo labeling. Despite being efficient, updating classifier prototype alone does not affect feature representation for the target domain. Target feature may not form clusters at all when the distribution mismatch between source and target is large enough. In this work we propose to simultaneously cluster on the target domain and match target clusters to source domain classes, namely anchored clustering. To further constrain feature update, we introduce additional global feature alignment and pseudo label filtering. Through the introduced anchored clustering, we achieve test-time training of more network parameters and achieve the state-of-the-art performance.


\section{Methodology}

\begin{figure}[!htb]
    \centering
    \includegraphics[width=0.99\linewidth]{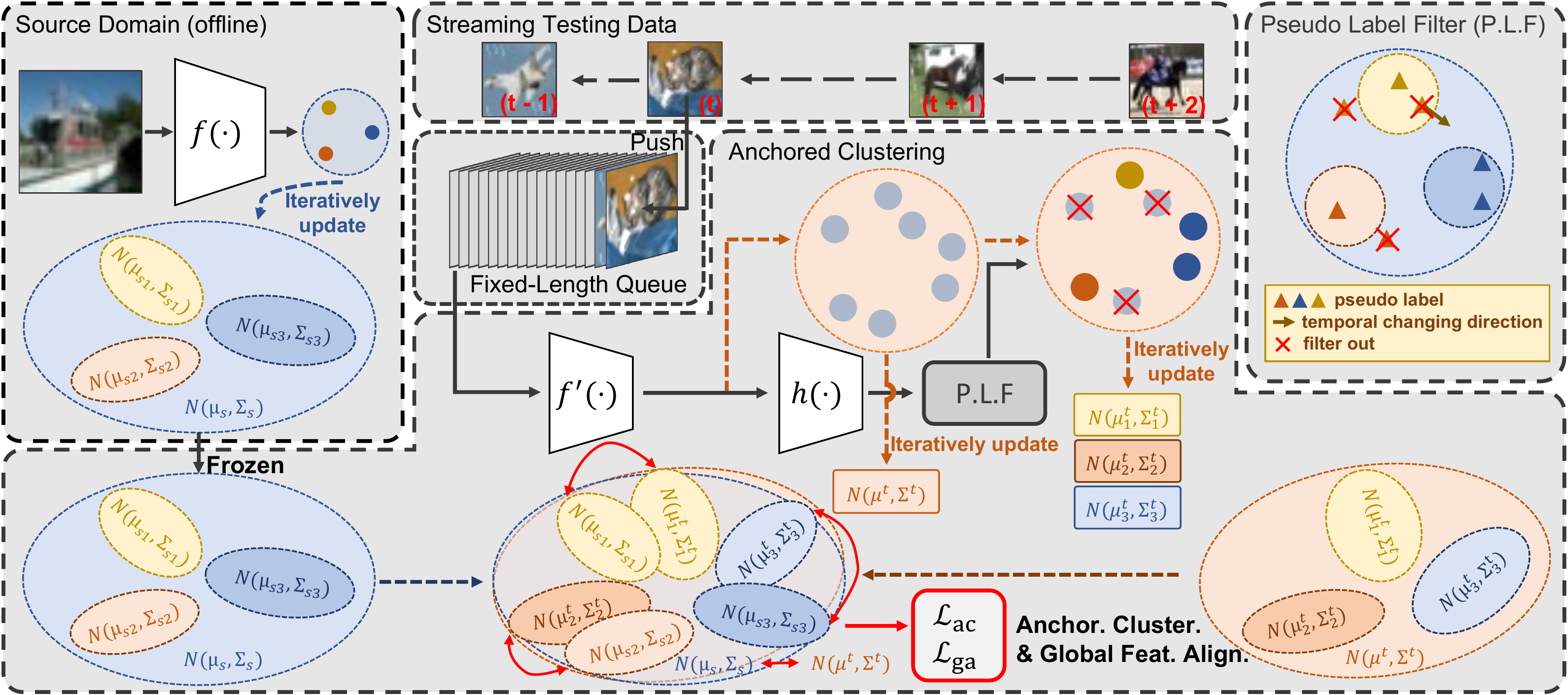}
    \caption{Overview of TTAC pipeline. i) In the source domain, we calculate category-wise and global statistics as anchors. ii) In the testing stage, samples are sequentially streamed and pushed into a fixed-length queue. Clusters in target domain are identified through anchored clustering with pseudo label filtering. Target clusters are then matched to the anchors in source domain to achieve test-time training.}\vspace{-0.5cm}
    \label{fig:overview}
\end{figure}

In this section we first introduce the anchored clustering objective for test-time training through pseudo labeling and then describe an efficient iterative updating strategy. An overview of the proposed pipeline is illustrated in Fig.~\ref{fig:overview}.

\subsection{Anchored Clustering for Test-Time Training}


Discovering cluster structures in the target domain has been demonstrated effective for unsupervised domain adaptation~\cite{tang2020unsupervised} and we develop an anchored clustering on the test data alone. 
We first use a mixture of Gaussians to model the clusters in the target domain, here each component Gaussian represents one discovered cluster. 
We further use the distributions of each category in the source domain as anchors for the target distribution to match against. In this way, test data features can simultaneously form clusters and the clusters are associated with source domain categories, resulting in improved generalization to target domain. Formally, we first write the mixture of Gaussians in the source and target domains $ p_s(x)=\sum_k \alpha_k \mathcal{N}(\mu_{sk},\Sigma_{sk}),\quad p_t(x)=\sum_k \beta_k \mathcal{N}(\mu_{tk},\Sigma_{tk})$, where $\{\mu_k\in\mathbb{R}^d,\Sigma_k\in\mathbb{R}^{d\times d}\}$ represent one cluster in the source/target domain and $d$ is the dimension of feature embedding.

   

Anchored clustering can be achieved by matching the above two distributions and one may directly minimize the KL-Divergence between the two distribution.
Nevertheless, this is non-trivial because the KL-Divergence between two mixture of Gaussians has no closed-form solution which prohibits efficient gradient-based optimization. Despite some approximations exist~\cite{hershey2007approximating}, without knowing the semantic labels for each Gaussian component, even a good match between two mixture of Gaussians does not guarantee target clusters are aligned to the correct source ones and this will severely harm the performance of test-time training. 
In light of these challenges, we propose a category-wise alignment.  Specifically, we allocate the same number of clusters in both source and target domains and each target cluster is assigned to one source cluster. We can then minimize the KL-Divergence between each pair of clusters as in Eq.~\ref{eq:KLD}. 
\vspace{-0.5cm}

\begin{equation}\label{eq:KLD}
\begin{split}
    \mathcal{L}_{ac}&=\sum_k D_{KL}(\mathcal{N}(\mu_{sk},\Sigma_{sk})||\mathcal{N}(\mu_{tk},\Sigma_{tk}))\\
    &=\sum_k -H(\mathcal{N}(\mu_{sk},\Sigma_{sk})) + H(\mathcal{N}(\mu_{sk},\Sigma_{sk}),\mathcal{N}(\mu_{tk},\Sigma_{tk}))
\end{split}
\end{equation}
\vspace{-0.3cm}

The KL-Divergence can be further decomposed into the entropy $H(\mathcal{N}(\mu_{sk},\Sigma_{sk}))$ and cross-entropy $H(\mathcal{N}(\mu_{sk},\Sigma_{sk}),\mathcal{N}(\mu_{tk},\Sigma_{tk}))$. It is commonly true that the source reference distribution $P_s(x)$ is fixed thus the entropy term is a constant $C$ and only the cross-entropy term is to be optimized.
Given the closed-form solution to the KL-Divergence between two Gaussian distributions, we now write the anchored clustering objective as,
\vspace{-0.3cm}

\begin{equation}\label{eq:anchored_clustering_loss}
    \mathcal{L}_{ac}= \sum_k \{\log \sqrt{2\pi^d|\Sigma_{tk}|} + \frac{1}{2}(\mu_{tk}-\mu_{sk})^\top\Sigma_{tk}^{-1}(\mu_{tk}-\mu_{sk}) + tr(\Sigma_{tk}^{-1}\Sigma_{sk})\} + C
\end{equation}
\vspace{-0.3cm}

The source cluster parameters can be estimated in an offline manner. These information will not cause any privacy leakage and only introduces a small computation and storage overheads. In the next section, we elaborate clustering in the target domain.
\vspace{-0.2cm}

\subsection{Clustering through Pseudo Labeling}
\vspace{-0.1cm}

In order to test-time train network with anchored clustering loss, one must obtain target cluster parameters $\{\mu_{tk},\Sigma_{tk}\}$. 
For a minibatch of target test samples $\set{B}^t=\{x_i\}_{i=1\dots N_B}$ at timestamp $t$, we first denote the predicted posterior as $P^t=softmax(h(f(x_i)))\in[0,1]^{B\times K}$ where $softmax(\cdot)$, $h(\cdot)$ and $f(\cdot)$ respectively denote a standard softmax function, the classifier head and backbone network. The pseudo labels are obtained via $\hat{y}_i=\arg\max_k P_{ik}^t$. Given the predicted pseudo labels we could estimate the mean and covariance for each component Gaussian with the pseudo labeled testing samples.
However, pseudo labels are always subject to model's discrimination ability. The error rate for pseudo labels is often high when the domain shift between source and target is large, directly updating the component Gaussian is subject to erroneous pseudo labels, a.k.a. confirmation bias~\cite{arazo2020pseudo}. To reduce the impact of incorrect pseudo labels, we first adopt a light-weight temporal consistency (TC) pseudo label filtering approach. Compared to co-teaching~\cite{han2018co} or meta-learning~\cite{li2019learning} based methods, this light-weight method does not introduce additional computation overhead and is therefore more suitable for test-time training.
Specifically, to alleviate the impact from the noisy predictions, we calculate the temporal exponential moving averaging posteriors $\widetilde{P}^t \in [0,1]^{N \times K}$ as below,

\vspace{-0.5cm}

\begin{equation}
    \widetilde{P}^t_i = 
            (1 - \xi) * \widetilde{P}^{t-1}_i + \xi * P^t_i\quad,\quad s.t.\quad\widetilde{P}^{0}_i=P^0_i
\end{equation}

The temporal consistency filtering is realized as in Eq.~\ref{eq:tc_filter} where $\tau_{TC}$ is a threshold determining the maximally allowed difference in the most probable prediction over time. If the posterior deviate from historical value too much, it will be excluded from target domain clustering.
\vspace{-0.2cm}

\begin{equation}\label{eq:tc_filter}
    F_i^{TC} = \mathbbm{1}((P_{i\hat{k}}^t - \widetilde{P}^{t-1}_{i\hat{k}}) > \tau_{TC}),\quad s.t. \quad \hat{k} = \arg\max_k(P_{ik}^t)
\end{equation}


Due to the sequential inference, test samples without enough historical predictions may still pass the TC filtering.  So, we further introduce an additional pseudo label filter directly based on the posterior probability as,
\vspace{-0.5cm}

\begin{equation}
    F_i^{PP}=\mathbbm{1}(\widetilde{P}^{t}_{i\hat{k}}>\tau_{PP})
\end{equation}

By filtering out potential incorrect pseudo labels, we update the component Gaussian only with the leftover target samples as below.
\vspace{-0.5cm}

\begin{equation}
    \mu_{tk} = \frac{\sum\limits_{i} F^{TC}_iF^{PP}_i\mathbbm{1}(\hat{y}_i=k)f(x_i)}{\sum\limits_i F^{TC}_iF^{PP}_i\mathbbm{1}(\hat{y}_i=k)},\quad
    \Sigma_{tk} = \frac{\sum\limits_i F^{TC}_iF^{PP}_i\mathbbm{1}(\hat{y}_i=k) (f(x_i)-\mu_{tk})^\top(f(x_i)-\mu_{tk})}{\sum\limits_i F^{TC}_iF^{PP}_i\mathbbm{1}(\hat{y}_i=k)}
\end{equation}
\vspace{-0.5cm}

\subsection{Global Feature Alignment}

As discussed above, test samples that do not pass the filtering will not contribute to the estimation of target clusters. Hence, anchored clustering may not reach its full potential without the filtered test samples. To exploit all available test samples, we propose to align global target data distribution to the source one. We define the global feature distribution of the source data as $\hat{p}_s(x)=\mathcal{N}(\mu_s,\Sigma_s)$ and the target data as $\hat{p}_t(x)=\mathcal{N}(\mu_t,\Sigma_t)$. To align two distributions, we again minimize the KL-Divergence as,
\vspace{-0.5cm}

\begin{equation}\label{eq:global_loss}
    \mathcal{L}_{ga}=D_{KL}(\hat{p}_s(x)||\hat{p}_t(x))
\end{equation}

Similar idea has appeared in~\cite{liu2021ttt++} which directly matches the moments between source and target domains~\cite{zellinger2017central} by minimizing the F-norm for the mean and covariance, i.e. $||\mu_t-\mu_s||^2_2+||\Sigma_t-\Sigma_s||^2_F$. However, designed for matching complex distributions represented as drawn samples, central moment discrepancy~\cite{zellinger2017central} requires summing infinite central moment discrepancies and the ratios between different order moments are hard to estimate.  For matching two parameterized Gaussian distributions KL-Divergence is more convenient with good explanation from a probabilistic point of view. Finally, we add a small constant to the diagonal of $\Sigma$ for both source and target domains to increase the condition number for better numerical stability.

\vspace{-0.2cm}

\subsection{Efficient Iterative Updating}
\vspace{-0.1cm}


Despite the distribution for source data can be trivially estimated from all available training data in a totally offline manner, estimating the distribution for target domain data is not equally trivial, in particular under the sTTT protocol.
In a related research~\cite{liu2021ttt++}, a dynamic queue of test data features are preserved to dynamically estimate the statistics, which will introduce additional memory footprint~\cite{liu2021ttt++}. 
To alleviate the memory cost we propose to iteratively update the running statistics for Gaussian distribution. 
Formally, we define $t$-th test minibatch as $\set{B}^t=\{x_i\}_{i=1\cdots N_{B}}$. Denoting the running mean and covariance at step $t$ as $\mu^t$ and $\Sigma^t$, we present the rules to update the mean and covariance in Eq.~\ref{eq:runningstatistics}. More detailed derivations and update rules for per cluster statistics are deferred to the Appendix. 

\vspace{-0.5cm}
\begin{equation}\label{eq:runningstatistics}
\begin{split}
    %
    & \mu^t = \mu^{t-1} + \delta^t, \quad
    \Sigma^t=\Sigma^{t-1}+a^t{\sum_{x_i\in\set{B}}\{(f(x_i)-\mu^{t-1})^\top(f(x_i)-\mu^{t-1})-\Sigma^{t-1}\}} - {\delta^t}^\top\delta^t \\
    & \delta^t=a^t{\sum\limits_{x_i\in\set{B}}(f(x_{i}) - \mu^{t-1})},\quad 
    N^t = N^{t-1} + |\set{B}^t|, \quad 
    a^t = \frac{1}{N^t}
\end{split}
\end{equation}
\vspace{-0.5cm}



Additionally, $N^t$ grows larger overtime. New test samples will have smaller contribution to the update of target domain statistics when $N^t$ is large enough. As a result, the gradient calculated from current minibatch will vanish. To alleviate this issue, we impose a clip on the value of $\alpha^t$ as below. As such, the gradient can maintain a minimal scale even if $N^t$ is very large. 


\vspace{-0.4cm}

\begin{equation}
    a^t = \left \{
        \begin{array}{lcl}
            \frac{1}{N^t} & & N^t < N_{clip} \\
            \frac{1}{N_{clip}} & & others
        \end{array}
        \right.
\end{equation}
\vspace{-0.5cm}

\subsection{TTAC Training Algorithm}
We summarize the training algorithm for the TTAC in Algo.~\ref{alg:main}. For effective clustering in target domain, we allocate a fixed length memory space, denoted as $\set{C} \in \mathbb{R}^{N_{C} \times H \times W \times \texttt{3}}$, to store the recent testing samples. In the sTTT protocol, we first make instant prediction on each testing sample, and only update the model when $N_B$ testing samples are accumulated. TTAC can be efficiently implemented, e.g. with two devices, one is for continuous inference and another is for model updating.

\vspace{-0.4cm}
\begin{algorithm}
\caption{Test-Time Anchored Clustering Training Algorithm }\label{alg:main}
\SetKwInOut{Input}{input}
\SetKwInOut{Return}{return}
\Input{A new testing sample batch $\set{B}^t=\{x_i\}_{i=1\dots N_B}$.}

\textcolor{gray}{\# Update the testing sample queue $\set{C}$.}

$\set{C}^t=\set{C}^t\setminus \set{B}^{t-N_C/N_B}$,\quad
$\set{C}^t=\set{C}^t\bigcup \set{B}^{t}$

\For{$ 1 $ \KwTo $N_{itr}$}
{
    \For{minibatch $\{x^t_k\}^N_{k=1}$ in  $\set{C}^t$}
    {
        
        
        \textcolor{gray}{\# Obtain the predicted posterior and pseudo labels}
        
        $P^t_i=softmax(h(f(x^t_i)))$,\quad
        $\hat{y}^t_i = \arg\max_k(P^t_{ik})$
        
        \textcolor{gray}{\# Calculate the global and per-cluster running mean and covariance by Eq.~\ref{eq:runningstatistics}}
        
        $\mu^t$,\quad $\Sigma^t$,\quad $\{\mu_k^t\}$,\quad $\{\Sigma_k^t\}$
        
        
        
        
        
        \textcolor{gray}{\# Optimize the combined loss by Eq.~\ref{eq:anchored_clustering_loss} and Eq.~\ref{eq:global_loss}}
        
        $\mathcal{L}=\mathcal{L}_{ac}+\lambda\mathcal{L}_{ga}$
        
        
        update network $f$ to minimize $\mathcal{L}$
    }
}


\end{algorithm}
\vspace{-0.5cm}





\section{Experiment}
\vspace{-0.3cm}

In this section, we first compare various existing methods based on the two key factors. Evaluation is then carried out on {six} test-time training datasets. We then ablate the components of TTAC. Further analysis on the cumulative performance, qualitative insights, etc. are provided at the end.

\vspace{-0.2cm}
\subsection{Datasets}
\vspace{-0.1cm}
We evaluate on 5 test-time training datasets and report the classification error rate~($\%$) throughout the experiment section. To evaluate the test-time training efficacy on corrupted target images, we use \textbf{CIFAR10-C/CIFAR100-C}~\cite{hendrycks2018benchmarking}, each consisting of 10/100 classes with 50,000 training samples of clean data and 10,000 corrupted test samples.  
We further evaluate test-time training on hard target domain samples with \textbf{CIFAR10.1}~\cite{pmlr-v97-recht19a}, which contains around 2,000 difficult testing images sampled over years of research on the original CIFAR-10 dataset.
To demonstrate the ability to do test-time training for synthetic data to real data transfer we further use \textbf{VisDA-C}~\cite{VisDA}, which is a challenging large-scale synthetic-to-real object classification dataset, consisting of 12 classes, 152,397 synthetic training images and 55,388 real testing images. To evaluate large-scale test-time training, we use \textbf{ImageNet-C}~\cite{hendrycks2018benchmarking} which consists of 1,000 classes and 15 types of corruptions on the 50,000 testing samples. Finally, to evaluate test-time training on 3D point cloud data, we choose \textbf{ModelNet40-C}~\cite{ModelNet40-C}, which consists of 15 common and realistic corruptions of point cloud data, with 9,843 training samples and 2,468 test samples.

\vspace{-0.2cm}
\subsection{Experiment Settings}

\noindent\textbf{Hyperparameters}. We use the ResNet-50~\cite{he2016deep} for image datasets and the DGCNN~\cite{wang2019dynamic} on ModelNet40-C. We optimize the backbone network $f(\cdot)$ by SGD with momentum on all datasets. On CIFAR10-C/CIFAR100-C and CIFAR10.1, we use (batchsize)~BS = 256 and (learning rate)~LR = 0.01, 0.0001, 0.01 respectively. On VisDA-C we use BS = 128 and LR = 0.0001, and on ModelNet40-C we use BS = 64 and LR = 0.001. More details of hyperparameters can be found in the Appendix. 

\noindent\textbf{Test-Time Training Protocols}.
We categorize test-time training based on two key factors. First, whether the training objective must be changed during training on the source domain, we use Y and N to indicate if training objective is allowed to be changed or not respectively. Second, whether testing data is sequentially streamed and predicted, we use O to indicate a sequential \textbf{O}ne-pass inference and M to indicate non-sequential inference, a.k.a. \textbf{M}ulti-pass inference. With the above criteria, we summarize 4 test-time training protocols, namely N-O, Y-O, N-M and Y-M, and the strength of the assumption increases from the first to the last protocols. 
Our sTTT setting  makes the weakest assumption, i.e. N-O. Existing methods are categorized by the four TTT protocols, we notice that some methods can operate under multiple protocols

\noindent\textbf{Competing Methods}.
We compare the following test-time training methods. Direct testing (\textbf{TEST}) without adaptation simply do inference on target domain with source domain model.
Test-time training (\textbf{TTT-R})~\cite{sun2020test} jointly trains the rotation-based self-supervised task and the classification task in the source domain, and then only train the rotation-based self-supervised task in the streaming test samples and make the predictions instantly. The default method is classified into the Y-M protocol.
Test-time normalization (\textbf{BN})~\cite{ioffe2015batch} moving average updates the batch normalization statistics by streamed data. The default method follows N-M protocol and can be adapted to N-O protocol.
Test-time entropy minimization (\textbf{TENT})~\cite{wang2020tent} updates the parameters of all batch normalization by minimizing the entropy of the model predictions in the streaming data. By default, TENT follows the N-O protocol and can be adapted to N-M protocol.
Test-time classifier adjustment (\textbf{T3A})~\cite{iwasawa2021test} computes target prototype representation for each category using streamed data and make predictions with updated prototypes. T3A follows the N-O protocol by default.
Source Hypothesis Transfer (\textbf{SHOT})~\cite{pmlr-v119-liang20a} freezes the linear classification head and trains the target-specific feature extraction module by exploiting balanced category assumption and self-supervised pseudo-labeling in the target domain. SHOT by default follows the N-M protocol and we adapt it to N-O protocol.
\textbf{TTT++}~\cite{liu2021ttt++} aligns source domain feature distribution, whose statistics are calculated offline, and target domain feature distribution by minimizing the F-norm between the mean covariance. TTT++ follows the Y-M protocol and we adapt it to N-O (removing contrastive learning branch) and Y-O protocols. Finally, we present our own approach, \textbf{TTAC}, which only requires a single pass on the target domain and does not have to modify the source training objective. We further modify TTAC for Y-O, N-M and Y-M protocols, for Y-O and Y-M we incorporate an additional contrastive learning branch~\cite{liu2021ttt++}. We could further combine TTAC with additional diversity loss and entropy minimization loss introduced in SHOT~\cite{pmlr-v119-liang20a}, denoted as TTAC+SHOT.

\vspace{-0.2cm}

\subsection{Test-Time Training on Corrupted Target Domain}
\vspace{-0.1cm}

We present the test-time training results on CIFAR10/100-C and ModelNet40-C datasets in Tab.~\ref{tab:categorization_table}, {and the results on ImageNet-C dataset in Tab.~\ref{tab:ImageNet}.} We make the following observations from the results.

\noindent\textbf{sTTT (N-O) Protocol}. 
We first analyze the results under the proposed sTTT (N-O) protocol. Our method outperforms all competing ones by a large margin. For example, $3\%$ improvement is observed on both CIFAR10-C and CIFAR100-C from the previous best (TTT++) and
{5-13\% improvement is observed on ImageNet-C compared with BN and TENT, and TTAC is superior in average accuracy and outperforms on 9 out of 15 types of corruptions compared with SHOT on ImageNet-C}. We further combine TTAC with the class balance assumption made in SHOT (TTAC+SHOT). With the stronger assumptions out method can further improve upon TTAC alone, in particular on ModelNet40-C dataset. This result demonstrates TTAC's compatibility with existing methods.

\noindent\textbf{Alternative Protocols}.
We further compare different methods under N-M, Y-O and Y-M protocols. Under the Y-O protocol, TTT++~\cite{liu2021ttt++} modifies the source domain training objective by incorporating a contrastive learning branch~\cite{chen2020simple}. To compare with TTT++, we also include the contrastive branch and observe a clear improvement on both CIFAR10-C and CIFAR100-C datasets. More TTT methods can be adapted to the N-M protocol which allows training on the whole target domain data multiple epochs. Specifically, we compared with BN, TENT and SHOT. With TTAC alone we observe substantial improvement on all three datasets and TTAC can be further combined with SHOT demonstrating additional improvement. Finally, under the Y-M protocol, we demonstrate very strong performance compared to TTT-R and TTT++. It is also worth noting that TTAC under the N-O protocol can already yield results close to TTT++ under the Y-M protocol, suggesting the strong test-time training ability of TTAC even under the most challenging TTT protocol.

\begin{table*}[htbp]
    \centering
   \caption{Comparison under different TTT protocols. Y/N indicates modifying source domain training objective or not. O/M indicate one pass or multiple passes test-time training. C10-C, C100-C and MN40-C refer to CIFAR10-C, CIFAR100-C and ModelNet40-C datasets respectively. All numbers indicate error rate in percentage.}
    \resizebox{0.75\linewidth}{!}{
        \begin{tabular}{l|cc|ccc}
        \toprule
        Method  & TTT Protocol & Assum. Strength & C10-C & C100-C & MN40-C \\
        \midrule
        TEST &  - & - & 29.15 & 60.34 & 34.62 \\
        \midrule
        BN~\cite{ioffe2015batch} &  N-O & Weak & 15.49 & 43.38 & 26.53 \\
        TENT~\cite{wang2020tent} &  N-O & Weak & 14.27 & 40.72 & 26.38 \\
        T3A~\cite{iwasawa2021test} &  N-O & Weak & 15.44 & 42.72 & 24.57 \\
        SHOT~\cite{pmlr-v119-liang20a} &  N-O & Weak & 13.95 & 39.10 & 19.71 \\
        TTT++~\cite{liu2021ttt++} &  N-O & Weak & 13.69 & 40.32 & - \\
        TTAC~(Ours)  &  N-O & Weak & \textbf{10.94} & 36.64 & 22.30 \\
        TTAC+SHOT~(Ours) & N-O & Weak & 10.99  & \textbf{36.39} & \textbf{19.21} \\
        \midrule
        TTT++~\cite{liu2021ttt++} & Y-O & Medium & 13.00 & 35.23 & -  \\
        TTAC~(Ours) & Y-O & Medium & \textbf{10.69} & \textbf{34.82} & - \\
        \midrule
        BN~\cite{ioffe2015batch} & N-M & Medium & 15.70 & 43.30 & 26.49 \\
        TENT~\cite{wang2020tent} & N-M & Medium & 12.60 & 36.30 & 21.23 \\
        SHOT~\cite{pmlr-v119-liang20a} & N-M & Medium & 14.70 & 38.10 & 15.99 \\
        TTAC~(Ours) & N-M & Medium & \textbf{9.42} & 33.55 & 16.77 \\
        TTAC+SHOT~(Ours) & N-M & Medium & 9.54 & \textbf{32.89} & \textbf{15.04} \\
        \midrule
        TTT-R~\cite{sun2020test} & Y-M & Strong & 14.30 & 40.40 & - \\ 
        TTT++~\cite{liu2021ttt++} & Y-M & Strong & 9.80 & 34.10 & - \\
        TTAC~(Ours) & Y-M & Strong & \textbf{8.52} & \textbf{30.57} & - \\
        \bottomrule
        \end{tabular}
    }
    \label{tab:categorization_table}
    \vspace{-0.3cm}
\end{table*}

\begin{table}[ht]
    \caption{Test-time training on ImageNet-C under the sTTT~(N-O) protocol.}
    \centering
    \resizebox{\linewidth}{!}{
        \begin{tabular}{c|ccccccccccccccc|c}
        \toprule
        Method & Birt & Contr & Defoc & Elast & Fog & Frost & Gauss & Glass & Impul & Jpeg & Motn & Pixel & Shot & Snow & Zoom & Avg \\
        \midrule
        TEST & 38.82 & 89.55 & 82.23 & 87.13 & 64.84 & 76.83 & 97.34 & 90.50 & 97.76 & 68.31 & 83.60 & 80.37 & 96.74 & 82.22 & 74.31 & 80.70 \\
        BN (N-O) & 32.33 & 50.93 & 81.28 & 52.98 & 42.21 & 64.13 & 83.25 & 83.64 & 82.52 & 59.18 & 66.23 & 49.45 & 82.59 & 62.34 & 52.51 & 63.04 \\
        TENT (N-O) & 31.39 & 40.27 & 75.68 & 42.03 & 35.38 & 64.32 & 84.92 & 84.96 & 81.43 & 46.84 & 49.48 & 39.77 & 84.21 & 49.23 & 43.49 & 56.89 \\
        SHOT (N-O) & 30.69 & \textbf{37.69} & \textbf{61.97} & 41.30 & \textbf{34.74} & 54.19 & 76.33 & 71.94 & 74.24 & 46.50 & \textbf{47.98} & \textbf{38.88} & 70.60 & 46.09 & \textbf{40.74} & 51.59 \\
        TTAC (N-O) & \textbf{30.36} & 38.84 & 69.06 & \textbf{39.67} & 36.01 & \textbf{50.20} & \textbf{66.18} & \textbf{70.17} & \textbf{64.36} & \textbf{45.59} & 51.77 & 39.72 & \textbf{62.43} & \textbf{44.56} & 42.80 & \textbf{50.11} \\
        \bottomrule
    \end{tabular}
    }
    \label{tab:ImageNet}
    \vspace{-0.2cm}
\end{table}
\vspace{-0.1cm}
\subsection{Additional Datasets}
\noindent\textbf{TTT on Hard Samples}. CIFAR10.1 contains roughly 2,000 new test images that were re-sampled after the research on original CIFAR-10 dataset, which consists of some hard samples and reflects the normal domain shift in our life. The results in Table.~\ref{tab:cifar101} demonstrate our method is better able to adapt to the normal domain shift. 


\noindent\textbf{TTT on Synthetic to Real Adaptation}. VisDA-C is a large-scale benchmark of synthetic-to-real object classification dataset. The setting of training on a synthetic dataset and testing on real data fits well with the real application scenario. On this dataset, we conduct experiments with our method under the N-O, Y-O and Y-M protocols and other methods under respective protocols, results are presented in Table.~\ref{tab:visda}. We make the following observations. First, our method (TTAC Y-O)
outperforms all methods except TTT++ under the Y-M protocol. This suggests TTAC is able to be deployed in the realistic test-time training protocol. Moreover, if training on the whole target data is allowed, TTAC (Y-M) further beats TTT++ by a large margin, suggesting the effectiveness of TTAC under a wide range of TTT protocols. 


\vspace{-0.5cm}

\begin{table}[htbp]
 \begin{minipage}{0.47\textwidth}
    \centering
    \caption{Test-time training on CIFAR10.1.}
        \resizebox{1\linewidth}{!}{
    \begin{tabular}{ccccccc}
    \toprule
    TEST & BN & TTT-R & TENT & SHOT & TTT++ & TTAC \\
    \midrule
    12.1 & 14.1 & 11.0 & 13.4 & 11.1 & 9.5 & \textbf{9.2} \\ 
    \bottomrule
    \end{tabular}
    }
    \label{tab:cifar101}
          \end{minipage}
 \hfill
        \begin{minipage}{0.53\textwidth}
         \centering
  \caption{Source-free sTTT on CIFAR10-C.}
  \resizebox{1\linewidth}{!}{
    \begin{tabular}{ccccccc}
    \toprule
    TEST  & BN & TENT  & T3A & SHOT & TTAC & TTAC+SHOT \\
    \midrule
    29.15 & 15.49 & 14.27 & 15.44 & 13.95 & 13.74 & \textbf{13.35} \\
    \bottomrule
    \end{tabular}%
    }
  \label{tab:source_blind}%
  \end{minipage}
\end{table}



\begin{table}[htbp]
    \centering
    \caption{Test-time training on VisDA. The numbers for competing methods are inherited from \cite{liu2021ttt++}.}
    \resizebox{\linewidth}{!}{
        \begin{tabular}{l|cccccccccccc|cc}
        \toprule
        Method & Plane & Bcycl & Bus & Car & Horse & Knife & Mcycl & Person & Plant & Sktbrd & Train & Truck & Per-class\\
        \midrule
        TEST & 56.52 & 88.71 & 62.77 & 30.56 & 81.88 & 99.03 & 17.53 & 95.85 & 51.66 & 77.86 & 20.44 & 99.51 & 65.19\\
        BN (N-M)~\cite{ioffe2015batch} & 44.38 & 56.98 & 33.24 & 55.28 & 37.45 & 66.60 & 16.55 & 59.02 & 43.55 & 60.72 & 31.07 & 82.98 & 48.99\\
        TENT (N-M)~\cite{wang2020tent} & 13.43 & 77.98 & 20.17 & 48.15 & 21.72 & 82.45 & 12.37 & 35.78 & 21.06 & 76.41 & 34.11 & 98.93 & 45.21\\
        SHOT (N-M)~\cite{pmlr-v119-liang20a} & 5.73 & \textbf{13.64} & 23.33 & 42.69 & 7.93 & 86.99 & 19.17 & 19.97 & 11.63 & 11.09 & 15.06 & \textbf{43.26} & 25.04 \\
        TFA (N-M)~\cite{liu2021ttt++} & 28.25 & 32.03 & 33.67 & 64.77 & 20.49 & 56.63 & 22.52 & 36.30 & 24.84 & 35.20 & 25.31 & 64.24 & 37.02 \\
        TTT++ (Y-M)~\cite{liu2021ttt++} & 4.13 & 26.20 & 21.60 & \textbf{31.70} & 7.43 & 83.30 & 7.83 & 21.10 & 7.03 & \textbf{7.73} & \textbf{6.91} & 51.40 & 23.03\\
        \midrule
        TTAC~(N-O) & 18.54 & 40.20 & 35.84 & 63.11 & 23.83 & 39.61 & 15.51 & 41.35 & 22.97 & 46.56 & 25.24 & 67.81 & 36.71 \\
        TTAC~(Y-O) & 7.19 & 29.99 & 22.52 & 56.58 & 8.14 & 18.41 & 8.25 & 22.28 & 10.18 & 23.98 & 13.55 & 67.02 & 24.01 \\
        TTAC~(Y-M) & \textbf{2.74} & 17.73 & \textbf{18.91} & 43.12 & \textbf{5.54} & \textbf{12.24} & \textbf{4.66} & \textbf{15.90} & \textbf{4.77} & 10.78 & 9.75 & 62.45 & \textbf{17.38} \\
        \toprule
        \end{tabular}
    }
    \label{tab:visda}
    \vspace{-0.3cm}

\end{table}

\subsection{Ablation Study}
\vspace{-0.1cm}

We conduct ablation study on CIFAR10-C dataset for individual components, including anchored clustering, pseudo label filtering, global feature alignment and finally the compatibility with contrastive branch~\cite{liu2021ttt++}. For anchored clustering alone, we use all testing samples to update cluster statistics. For pseudo label filtering alone, we implement as predicting pseudo labels followed by filtering, then pseudo labels are used for self-training. We make the following observations from Tab.~\ref{tab:ablation}. Under both N-O and N-M protocols, introducing anchored clustering or pseudo label filtering alone improves over the baseline, e.g. under N-O $29.15\%\rightarrow 14.32\%$ for anchored clustering and $29.15\%\rightarrow15.00\%$ for pseudo label filtering. When anchored clustering is combined with pseudo label filtering, we observe a significant boost in performance. This is due to more accurate estimation of category-wise cluster in the target domain {and this reflects matching directly in the feature space 
may be better than minimizing 
cross-entropy with pseudo labels}. We further evaluate aligning global features alone with KL-Divergence. This achieves relatively good performance and obviously outperforms the L2 distance alignment adopted in \cite{liu2021ttt++}. Finally, we combine all three components and the full model yields the best performance. When contrast learning branch is included, TTAC achieves even better results.



%

\begin{table}[htbp]
\vspace{-0.4cm}
  \centering
  \caption{Ablation study for individual components on CIFAR10-C dataset.}
    \resizebox{1\linewidth}{!}{
    \begin{tabular}{lcccccrrrrrrrr}
    \toprule
    TTT Protocol & -     & \multicolumn{5}{c}{N-O}               & \multicolumn{1}{c}{Y-O} & \multicolumn{5}{c}{N-M}               &  \multicolumn{1}{c}{Y-M}\\
    \cmidrule(lr){1-1} \cmidrule(lr){2-2} \cmidrule(lr){3-7} \cmidrule(lr){8-8} \cmidrule(lr){9-13} \cmidrule(lr){14-14} 
    Anchored Cluster. & -     & \checkmark & -     & \checkmark & -     & \multicolumn{1}{c}{\checkmark} & \multicolumn{1}{c}{\checkmark} & \multicolumn{1}{c}{\checkmark} & \multicolumn{1}{c}{\checkmark} & \multicolumn{1}{c}{-} & \multicolumn{1}{c}{-} & \multicolumn{1}{c}{\checkmark} & \multicolumn{1}{c}{\checkmark} \\
    Pseudo Label Filter. & -     & -     & \checkmark & \checkmark & -     & \multicolumn{1}{c}{\checkmark} & \multicolumn{1}{c}{\checkmark} & \multicolumn{1}{c}{-} & \multicolumn{1}{c}{\checkmark} & \multicolumn{1}{c}{-} & \multicolumn{1}{c}{-} & \multicolumn{1}{c}{\checkmark} & \multicolumn{1}{c}{\checkmark} \\
    Global Feat. Align. & -     & -     & -     & -     & KLD   & \multicolumn{1}{c}{KLD} & \multicolumn{1}{c}{KLD} & \multicolumn{1}{c}{-} & \multicolumn{1}{c}{-} & \multicolumn{1}{c}{L2 Dist.\cite{liu2021ttt++}} & \multicolumn{1}{c}{KLD} & \multicolumn{1}{c}{KLD} & \multicolumn{1}{c}{KLD} \\
    Contrast. Branch~\cite{liu2021ttt++} & -     & -     & -     & -     & -     & \multicolumn{1}{c}{-} & \multicolumn{1}{c}{\checkmark} & \multicolumn{1}{c}{-} & \multicolumn{1}{c}{-} & \multicolumn{1}{c}{-} & \multicolumn{1}{c}{-} & \multicolumn{1}{c}{-} & \multicolumn{1}{c}{\checkmark} \\
    Avg Acc & 29.15 & 14.32 & 15.00 & 11.33 & 11.72 & 10.94 & 10.69 & 11.11 & 10.01 & \multicolumn{1}{c}{11.87} & 10.8  & 9.42  & 8.52 \\
    \bottomrule
    \end{tabular}%
    }
  \label{tab:ablation}%
  \vspace{-0.3cm}
\end{table}%

\subsection{Additional Analysis}

\begin{figure}
\subfloat[Test-time cumulative error]{\includegraphics[width=0.45\linewidth]{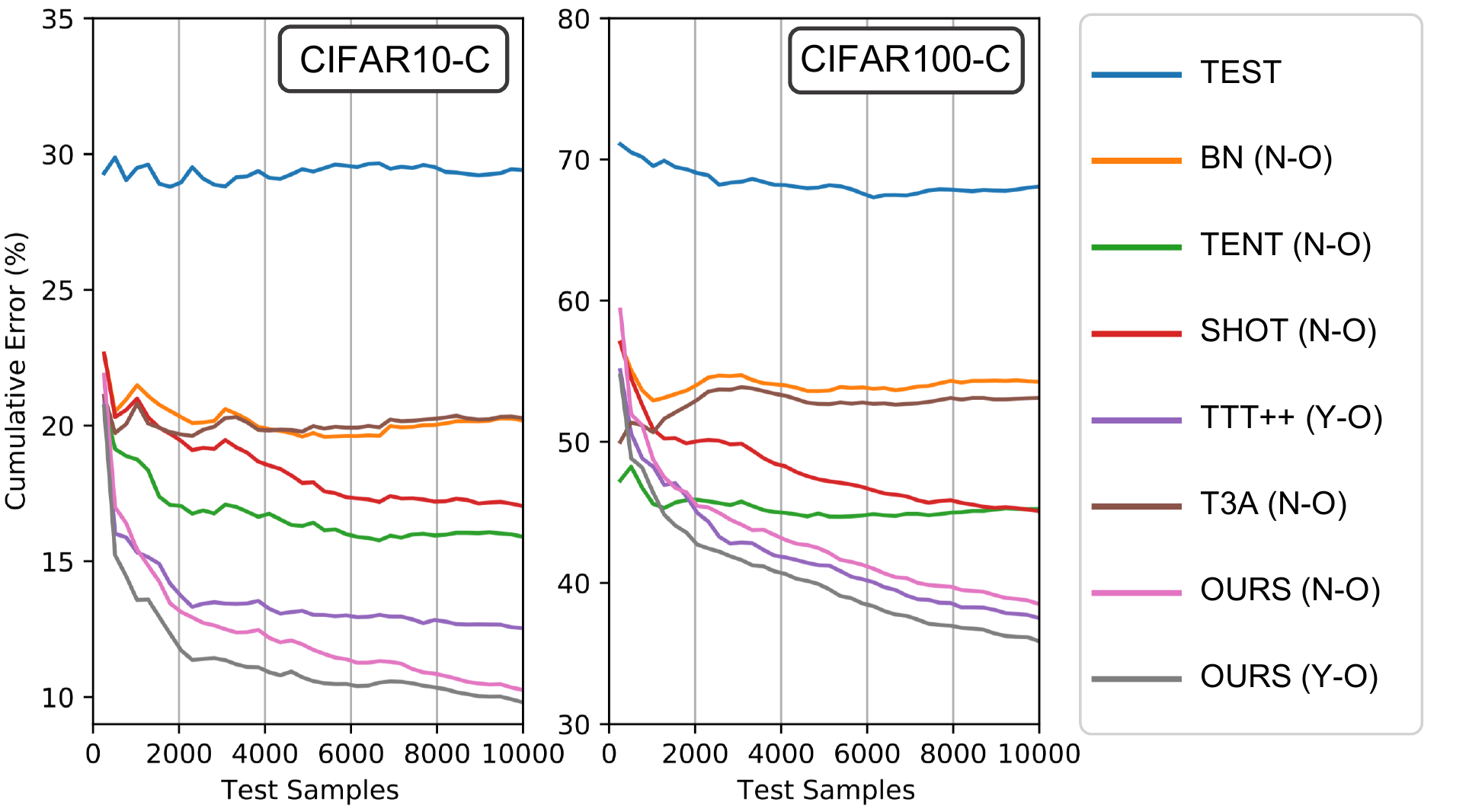}}
\subfloat[TTT++ Feature]{\includegraphics[width=0.25\linewidth]{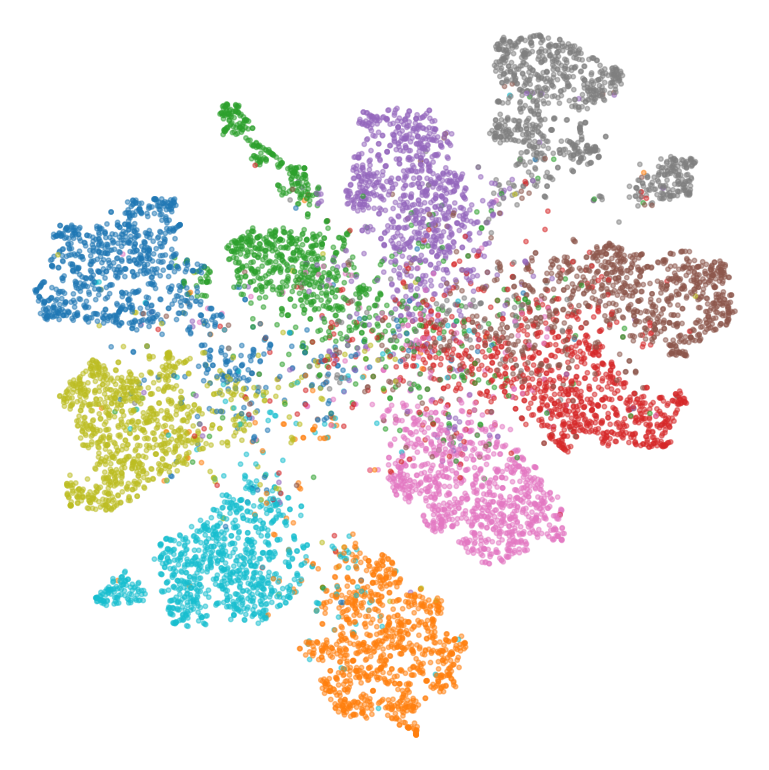}}
\subfloat[TTAC Feature]{\includegraphics[width=0.25\linewidth]{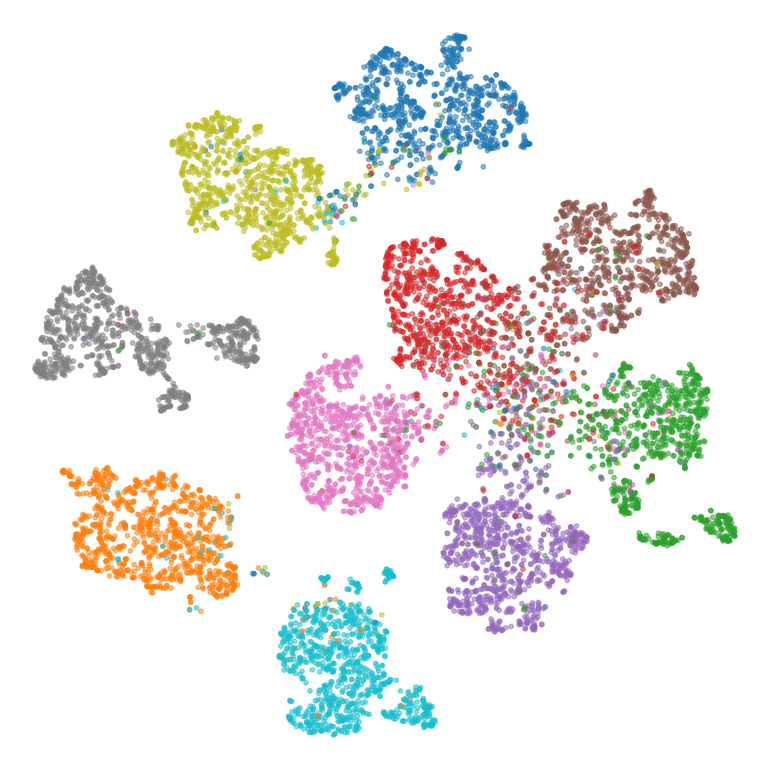}}

\caption{(a) Comparison of test-time cumulative error under one-pass protocol. (b) T-SNE visualization of TTT++ feature embedding. (c) T-SNE visualization of TTAC feature embedding.}\label{fig:add_study}
\vspace{-0.1cm}

\end{figure}

\noindent\textbf{Cumulative performance under sTTT}. We illustrate the cumulative error under the sTTT protocol in Fig.~\ref{fig:add_study}~(a). For both datasets TTAC outperforms competing methods from the early stage of test-time training. The advantage is consistent throughout the TTT procedure.


\noindent\textbf{TSNE Visualization of TTAC features}.
We provide qualitative results for test-time training by visualizing the adapted features through T-SNE~\cite{van2008visualizing}. In Fig.~\ref{fig:add_study}~(b) and Fig.~\ref{fig:add_study}~(c), we compared the features learned by TTT++~\cite{liu2021ttt++} and TTAC~(Ours). We observe a better separation between classes by TTAC, implying an improved classification accuracy.


\noindent\textbf{Source-Free Test-Time Training}. 
TTT aims to adapt model to target domain data by doing simultaneous training and sequential inference. It has been demonstrated some light-weight information, e.g. statistics, from source domain will greatly improve the efficacy of TTT. Nevertheless, under a more strict scenario where source domain information is strictly blind, TTAC can still exploit classifier prototypes to facilitate anchored clustering. Specifically, we normalize the category-wise weight vector with the norm of corresponding target domain cluster center as prototypes. Then, we build source domain mixture of Gaussians by taking prototypes as mean with a fixed covariance matrix. The results on CIFAR10-C are presented in Tab.~\ref{tab:source_blind}. It is clear that even without any statistical information from source domain, TTAC still outperforms all competing methods.

\noindent\textbf{Test Sample Queue and Update Epochs.}
{
Under the sTTT protocol, we allow all competing methods to maintain the same test sample queue and multiple update epochs on the queue. To analyse the significance of the sample queue and update epochs, we evaluate BN, TENT, SHOT and TTAC on CIFAR10-C and ImageNet-C level 5 snow corruption evaluation set under different number of update epochs on test sample queue and under a without queue protocol, i.e. only update model w.r.t. the current test sample batch. As the results presented in Tab.~\ref{tab:QueueAccuracyAnalysis}, we make the following observations. i) Maintaining a sample queue can substantially improve the performance of methods that estimate target distribution, e.g. TTAC ($11.91\rightarrow10.88$ on CIFAR10-C) and SHOT ($15.18\rightarrow13.96$ on CIFAR10-C). This is due to more test samples giving a better estimation of true distribution. ii) Consistent improvement can be observed with increasing update epochs for SHOT and TTAC. We ascribe this to iterative pseudo labeling benefiting from more update epochs.
}

\begin{table}[!ht]
    \caption{Comparing with and without test sample queue and different numbers of model update epochs. w/ Queue maintains a test sample queue with 4096 samples; w/o Queue maintains a single mini-batch with 256 and 128 samples on CIFAR10-C and ImageNet-C respectively.}
    \centering
    \resizebox{0.70\linewidth}{!}{
    \begin{tabular}{c|cccc|c|cc|c}
    \toprule
         & \multicolumn{5}{c|}{CIFAR10-C} & \multicolumn{3}{c}{ImageNet-C} \\
    \midrule
         & \multicolumn{4}{c|}{w/ Queue } & \multicolumn{1}{c|}{w/o Queue} & \multicolumn{2}{c|}{w/ Queue} & w/o Queue \\
    \midrule
        \#Epochs & 1 & 2 & 3 & 4* & 1 & 1 & 2* & 1 \\
    \midrule
        BN & 15.84 & 15.99 & 16.04 & 16.00 & 15.44 & 62.34 & 62.34 & 62.59\\
        TENT & 13.35 & 13.83 & 13.85 & 13.87 & 13.48 & 47.82 & 49.23 & 48.39\\
        SHOT & 13.96 & 13.93 & 13.83 & 13.75 & 15.18 & 46.91 & 46.09 & 51.46\\
        TTAC & \textbf{10.88} & \textbf{10.80} & \textbf{10.58} & \textbf{9.96} & \textbf{11.91} & \textbf{45.44} & \textbf{44.56} & \textbf{46.64}\\
    \bottomrule
    \end{tabular}
    }
    \label{tab:QueueAccuracyAnalysis}
\end{table}

\noindent\textbf{Computation Cost Measured in Wall-Clock Time.}
{
Test sample queue and multiple update epochs introduce additional computation overhead. To investigate the impact on efficiency, we measure the overall wall time as the time elapsed from the beginning to the end of test-time training, including all I/O overheads. The per-sample wall time is then calculated as the overall wall time divided by the number of test samples. 
We report the per-sample wall time (in seconds) for BN, TENT, SHOT and TTAC in Tab.~\ref{tab:QueueTimeAnalysis} under different update epoch settings and without queue setting.
The Inference row indicates the per-sample wall time in a single forward pass including the data I/O overhead.
We observe that, under the same experiment setting, BN and TENT are more computational efficient, but TTAC is only twice more expensive than BN and TENT if no test sample queue is preserved (0.0083 v.s. 0.0030/0.0041) while the performance of TTAC w/o queue is still better than TENT (11.91 v.s. 13.48).
In summary, TTAC is able to strike a balance between computation efficiency and performance depending on how much computation resource is available. This suggests allocating a separate device is only necessary when securing best performance is the priority.
}

\begin{table}[!ht]
    \caption{The per-sample wall time (measured in seconds) on CIFAR10-C under sTTT protocol.}
    \centering
    \resizebox{0.50\linewidth}{!}{
    \begin{tabular}{c|cccc|c}
    \toprule
         & \multicolumn{4}{c|}{w/ Queue } & \multicolumn{1}{c}{w/o Queue}\\
    \midrule
        \#Epochs & 1 & 2 & 3 & 4 & 1 \\
    \midrule
        BN & 0.0136 & 0.0220 & 0.0293 & 0.0362 & 0.0030\\
        TENT & 0.0269 & 0.0399 & 0.0537 & 0.0663 & 0.0041\\
        SHOT & 0.0479 & 0.0709 & 0.0942 & 0.1183 & 0.0067\\
        TTAC & 0.0516 & 0.0822 & 0.1233 & 0.1524 & 0.0083\\
    \midrule
        Inference & 0.0030 & 0.0030 & 0.0030 & 0.0030 & 0.0030\\
    \bottomrule
    \end{tabular}
    }
    \label{tab:QueueTimeAnalysis}
\end{table}

\vspace{-0.5cm}
\section{Conclusion}
\vspace{-0.3cm}
Test-time training~(TTT) tackles the realistic challenges of deploying domain adaptation on-the-fly. In this work, we are first motivated by the confused evaluation protocols for TTT and propose two key criteria, namely modifying source training objective and sequential inference, to further categorize existing methods into four TTT protocols. Under the most realistic protocol, i.e. sequential test-time training (sTTT), we develop a test-time anchored clustering (TTAC) approach to align target domain features to the source ones. Unlike batchnorm and classifier prototype updates, anchored clustering allows all network parameters to be trainable, thus demonstrating stronger test-time training ability. We further propose pseudo label filtering and an iterative update method to improve anchored clustering and save memory footprint respectively. Experiments on {six} datasets verified the effectiveness of TTAC under sTTT as well as other TTT protocols.

\noindent\textbf{Acknowledgement} This work was supported in part by the National Natural Science Foundation of China~(NSFC) under Grant 62106078, Guangdong R\&D key project of China (No.: 2019B010155001), and the Program for Guangdong Introducing Innovative and Enterpreneurial Teams (No.: 2017ZT07X183).
\newpage
{
\small
\bibliographystyle{plain}
\bibliography{references}
}

\newpage

\appendix



\vbox{
\centering
\Large{\textbf{Appendix of "Revisiting Realistic Test-Time Training: Sequential Inference and Adaptation by Anchored Clustering"}}
\vskip 0.3in
}

In this appendix, we first provide more details for the derivation of iterative updating target domain cluster parameters. We further provide more details of the hyperparameters used in TTAC. Finally, we present evaluation of TTAC with transformer backbone, ViT~\cite{dosovitskiy2020vit}, additional evaluation of TTAC update epochs, the stability of TTAC under different data streaming orders and compared alternative target clustering updating strategies.

\section{Derivations of Efficient Iterative Updating}


The mean and covariance for each target domain cluster can be naively estimated through Maximum Likelihood Estimation~(MLE) as below. The existing solution in TTT++~\cite{liu2021ttt++} stores the recent one thousand testing samples and their features for MLE.

\begin{equation}
    \mu = \frac{1}{N}\sum_{i=1}^{N}{f(x_i)},\quad
    \Sigma = \frac{1}{N}\sum_{i=1}^{N}{(f(x_i) - \mu)^{\top}(f(x_i) - \mu)}
\end{equation}

When $N$ is very large, it is inevitable that a very large memory space must be allocated to store all features $F \in \mathbb{R}^{N \times D}$, e.g. the VisDA dataset has 55k testing samples and a naive MLE prohibits efficient test-time training. 
In the manuscript, we propose to online update target domain feature distribution parameters without caching sample features as Eq.~8. 
The detailed derivations are now presented as follows.
Formally, we denote the running mean and covariance at step $t-1$ as $\mu^{t-1}$ and $\Sigma^{t-1}$, and the test minibatch at step $t$ as $\set{B}^t=\{x_i\}_{i=1\cdots N_B}$. The following is the derivation of $\mu^t$.

\begin{equation}
    \mu^t = \frac{1}{N^t}\sum_{i=1}^{N^t}{f(x_i)}, \quad s.t. \quad N^t = N^{t-1} + |\set{B}^t|
\end{equation}

\begin{equation}
\begin{split}
    \mu^t 
        & = \frac{1}{N^t}(\sum_{i=1}^{N^{t-1}}{f(x_i)} + \sum_{i=N^{t-1}+1}^{N^t}{f(x_i)}) \\
        & = \frac{1}{N^t}(N^{t-1}\cdot\mu^{t-1} + \sum_{i=N^{t-1}+1}^{N^t}{f(x_i)}) \\
        & = \mu^{t-1} + \frac{1}{N^t}\sum_{x_i\in \set{B}^t}{(f(x_i) - \mu^{t-1})}
\end{split}
\end{equation}

To simplify the expression, we denote  $\delta^t=\frac{1}{N^t}\sum_{x_i\in \set{B}^t}{(f(x_i) - \mu^{t-1})}$, so $\mu^t = \mu^{t-1} + \delta^t$. The following is the derivation of $\Sigma^t$. For the ease of calculation, we use the asymptotic unbiased estimator of $\Sigma^t$ as shown as below.

\begin{equation}
\begin{split}
    \Sigma^t 
        & = \frac{1}{N^t}\sum_{i=1}^{N^t}{(f(x_i) - \mu^t)^{\top} (f(x_i) - \mu^t)} \\
        & = \frac{1}{N^t}\sum_{i=1}^{N^t}{(f(x_i) - \mu^{t-1} - \delta^t)^{\top} (f(x_i) - \mu^{t-1} - \delta^t)} \\
        & = \frac{1}{N^t}\sum_{i=1}^{N^t}{\{(f(x_i) - \mu^{t-1})^\top(f(x_i) - \mu^{t-1}) - {\delta^t}^\top(f(x_i) - \mu^{t-1}) - (f(x_i) - \mu^{t-1})^\top\delta^t + {\delta^t}^\top{\delta^t} \}} \\
        & = \frac{1}{N^t}(\sum_{i=1}^{N^{t-1}}{(f(x_i) - \mu^{t-1})^\top(f(x_i) - \mu^{t-1})} + \sum_{x_i \in \set{B}^t}{(f(x_i) - \mu^{t-1})^\top(f(x_i) - \mu^{t-1})} \\
        & + \sum_{i=1}^{N^t}{\{  - {\delta^t}^\top(f(x_i) - \mu^{t-1}) - (f(x_i) - \mu^{t-1})^\top\delta^t\}}) + {\delta^t}^\top{\delta^t} \\
        & = \frac{1}{N^t}(N^{t-1} \cdot \Sigma^{t-1} + \sum_{x_i \in \set{B}^t}{(f(x_i) - \mu^{t-1})^\top(f(x_i) - \mu^{t-1}))} - {\delta^t}^\top\delta^t \\
        & = \Sigma^{t-1} + \frac{1}{N^t}\sum_{x_i \in \set{B}^t}{\{(f(x_i) - \mu^{t-1})^\top(f(x_i) - \mu^{t-1}) - \Sigma^{t-1}\}} - {\delta^t}^\top{\delta^t}
\end{split}
\end{equation}

Furthermore, we give the formulations of the running mean $\mu^t_k$ and covariance $\Sigma^t_k$ for the $k^{th}$ target domain cluster as below.

\begin{equation}\label{eq:category-wise-runningstatistics}
\begin{split}
    & \delta_k^t=\frac{1}{N_k^t}{\sum_{x_i\in\set{B}^t}F^{TC}_iF^{PP}_i\mathbbm{1}(\hat{y}_i=k)(f(x_i)-\mu_k^{t-1})}, \\
    & s.t. \quad N_k^t = N_k^{t-1} + \sum_{x_i\in\set{B}^t}F^{TC}_iF^{PP}_i\mathbbm{1}(\hat{y}_i=k)\\
    & \mu_k^t = \mu_k^{t-1} + \delta_k^t, \\
    &\Sigma_k^t=\Sigma_k^{t-1}+\frac{1}{N_k^t}{\sum_{x_i\in\set{B}^t}F^{TC}_iF^{PP}_i\mathbbm{1}(\hat{y}_i=k)\{(f(x_i)-\mu_k^{t-1})^\top(f(x_i)-\mu_k^{t-1})-\Sigma_k^{t-1}\}} - {\delta_k^t}^\top\delta_k^t
\end{split}
\end{equation}

Similarly to $N_{clip}$ for the threshold used to clip the $N^t$ protecting the gradient of new test samples, we use $N_{clip\_k}$ as the threshold to clip the $N^t_k$ for each target domain cluster.

\section{Hyperparameter Values}
We provide the details of hyperparameters in this section. Hyperparameters are shared across multiple TTT protocols except for $N_C$ and $N_{itr}$ which are only applicable under one-pass adaptation protocols. The details are shown as Tab.~\ref{tab:hyperparameters}. $\alpha_k$ and $\beta_k$ respectively represent the prevalence of each category, here we set them to 1 over the number of categories. $N_C$ indicates the length of the testing sample queue $C$ under the sTTT protocol, and $N_{itr}$ controls the update epochs on this queue. $\tau_{TC}$ and $\tau_{PP}$ are the thresholds used for  pseudo label filtering. $N_{clip}$ and $N_{clip\_k}$ are the upper bounds of sample counts in the iterative updating of global statistics and target cluster statistics respectively. Finally $\lambda$ is the coefficient of $\mathcal{L}_{ga}$, which takes the default value of 1. All models are implemented by the PyTorch 1.10.2 framework, CUDA 11.3 with an NVIDIA RTX 3090 GPU.

\begin{table}[]
    \centering
    \caption{Hyper-parameters are used in our method.}
    \begin{tabular}{l|cccccccccc}
    \toprule
      Dataset & $\alpha_k$ & $\beta_k$ & $N_C$ & $N_{itr}$ & $\xi$ & $\tau_{TC}$ & $\tau_{PP}$ & $N_{clip}$ & $N_{clip\_k}$ & $\lambda$\\
    \midrule
      CIFAR10-C  &  0.1 & 0.1 & 4096 & 4 & 0.9 & -0.001 & 0.9 & 1280 & 128 & 1.0\\
      CIFAR100-C  & 0.01  & 0.01 & 4096 & 4 & 0.9 & -0.001 & 0.9 & 1280 & 64 & 1.0\\
      CIFAR10.1 & 0.1 & 0.1 & 4096 & 4 & 0.9 & -0.001 & 0.9 & 1280 & 128 & 1.0\\
      VisDA-C & $\frac{1}{12}$ & $\frac{1}{12}$ & 4096 & 4 & 0.9 & -0.01 & 0.9 & 1536 & 128 & 1.0\\
      ModelNet40-C & 0.025 & 0.025 & 4096 & 6 & 0.9 & -0.1 & 0.5 & 1280 & 128 & 1.0\\
      ImageNet-C & 0.001 & 0.001 & 4096 & 2 & 0.9 & -0.01 & 0.9 & 1280 & 64 & 1.0 \\
    \bottomrule
    \end{tabular}
    \label{tab:hyperparameters}
\end{table}

\section{Additional Evaluation}

\subsection{Evaluation of TTAC with Transformer Backbone}
In this section, we provide additional evaluation of TTAC with a transformer backbone, ViT~\cite{dosovitskiy2020vit}. In specific, we pre-train ViT on CIFAR10 clean dataset and then follow the sTTT protocol to do test-time training on CIFAR10-C. The results are presented in Tab.~\ref{tab:ViT}. We report the average~(Avg) and standard deviation~(Std) of accuracy over all 15 categories of corruptions. Again, TTAC consistently outperform all competing methods with transformer backbone.

\begin{table}[]
    \centering
    \caption{The results using ViT backbone on CIFAR10-C dataset.}
    \resizebox{\linewidth}{!}{
        \begin{tabular}{l|ccccccccccccccc|cc}
        \toprule
            Method & Bird & Contr & Defoc & Elast & Fog & Frost & Gauss & Glass & Impul & Jpeg & Motn & Pixel & Shot & Snow & Zoom & Avg & Std\\
        \midrule
            TEST & 2.29 & 16.24 & 4.83 & 9.45 & 13.60 & 6.73 & 24.52 & 18.23 & 24.48 & 12.63 & 7.63 & 14.57 & 23.02 & 5.29 & 3.50 & 12.47 & 7.36 \\
            BN    & 2.29 & 16.24 & 4.83 & 9.45 & 13.60 & 6.73 & 24.52 & 18.23 & 24.48 & 12.63 & 7.63 & 14.57 & 23.02 & 5.29 & 3.50 & 12.47 & 7.36  \\
            TENT  & \textbf{1.84} & 3.55  & \textbf{3.31} & 7.01 & \textbf{5.57}  & 4.09 & 60.97 & 10.20 & 61.12 & 9.72  & 4.93 & 3.87  & 22.47 & 4.55 & \textbf{2.64} & 13.72 & 19.19 \\
            SHOT  & 2.00 & \textbf{3.13}  & 3.46 & 6.63 & 5.79  & 4.06 & 11.65 & 9.39  & 10.58 & 9.69  & 5.03 & 3.63  & 10.05 & 4.35 & 2.70 & 6.14  & 3.15  \\
            TTT++ & 1.91 & 4.14  & 3.88 & \textbf{6.58} & 6.27  & 4.00    & 10.08 & 8.59  & 8.85  & 9.66  & \textbf{4.68} & \textbf{3.62}  & 9.17  & 4.28 & 2.74 & 5.90  & 2.64  \\
            TTAC (Ours)  & 2.15 & 4.05  & 3.91 & 6.62 & 5.67  & \textbf{3.75} & \textbf{9.26}  & \textbf{7.95}  & \textbf{7.97}  & \textbf{8.55}  & 4.75 & 3.87  & \textbf{8.24}  & \textbf{3.93} & 2.94 & \textbf{5.57} & \textbf{2.24} \\
        \bottomrule
        \end{tabular}
    }
    \label{tab:ViT}
\end{table}

\subsection{Impact of TTAC Update Epochs on Cached Testing Sample}
Under the sTTT protocol, we perform multiple iterations of adaptation on cached testing sample queue. Preserving a history of testing samples is a commonly practice in test-time training. For example, T3A~\cite{iwasawa2021test} preserves a support set, which contains testing samples and the pseudo labels, to update classifier prototypes. TTT++~\cite{liu2021ttt++} preserves a testing sample queue to estimate global feature distribution. For these methods, both raw testing samples and features must be cached simultaneously, in comparison, we only cache the raw data samples and target domain clusters are estimated in an online fashion.

Here, we analyze the impact of TTAC update epochs on cached testing samples. The results are presented in Tab.~\ref{tab:iterations}, where we make the following observations. First, the error rate is decreasing as the number of epochs increases, while at the cost of more computation time. But this can be solved by allocating a separate device for model adaptation. Second, the error rate saturates at $N_{itr}=4$ suggesting only a few epochs is necessary to achieve good test-time training on target domain.

\begin{table}[htbp]
    \centering
    \caption{The impact of TTAC update epochs under the sTTT protocol.}
    \resizebox{\linewidth}{!}{
        \begin{tabular}{c|ccccccccccccccc|c}
            \toprule
            $N_{itr}$ & Bird & Contr & Defoc & Elast & Fog & Frost & Gauss & Glass & Impul & Jpeg & Motn & Pixel & Shot & Snow & Zoom & Avg \\
            \midrule
             1 & 6.57 & 8.20 & 8.57 & 15.82 & 11.61 & 11.60 & 17.46 & 22.66 & 20.99 & 11.97 & 10.44 & 13.79 & 15.40 & 10.96 & 7.49 & 12.90\\
             2 & 6.82 & 8.12 & 8.77 & 15.96 & 11.79 & 11.17 & 15.49 & 23.53 & 19.78 & 12.28 & 10.19 & 13.22 & 16.28 & 10.84 & 7.49 & 12.78 \\
            3 & 6.80 & 8.11 & 8.53 & 15.94 & 11.36 & 10.89 & 14.87 & 22.67 & 18.94 & 11.77 & 9.83 & 12.51 & 15.91 & 10.58 & 7.35 & 12.40 \\
            \textbf{4} & \textbf{6.41} & 8.05 & \textbf{7.85} & 14.81 & \textbf{10.28} & \textbf{10.51} & \textbf{13.06} & 18.36 & \textbf{17.35} & \textbf{10.80} & 8.97  & \textbf{9.34}  & \textbf{11.61} & \textbf{10.01} & \textbf{6.68} & \textbf{10.94} \\
            6 & 6.42 & \textbf{7.64} & 7.97 & \textbf{14.66} & 10.66 & 10.59 & 13.30 & \textbf{18.29} & 17.61 & 10.86 & \textbf{8.94}  & 9.36  & 11.76 & 10.03 & 6.73 & 10.98\\
            \bottomrule
        \end{tabular}
    }
    \label{tab:iterations}
\end{table}

\subsection{Impact of Data Streaming Order}
The proposed sTTT protocols assumes test samples arrive in a stream and inference is made instantly on each test sample. The result for each test sample will not be affected by any following ones. In this section, we investigate how the data streaming order will affect the results. Specifically, we randomly shuffle all testing samples in CIFAR10-C for 10 times with different seeds and calculate the mean and standard deviation of test accuracy under sTTT protocol. The results in Tab.~\ref{tab:stream_order} suggest TTAC maintains consistent performance regardless of data streaming order.

\begin{table}[]
    \centering
    \caption{The performance of TTAC under different data streaming orders. }
    \resizebox{\linewidth}{!}{
        \begin{tabular}{l|cccccccccc|cc}
        \toprule
           Random Seed & 0 & 10 & 20 & 200 & 300 & 3000 & 4000 & 40000 & 50000 & 500000 & Avg \\
        \midrule
           Error ($\%$) & 10.01 & 10.06 & 10.05 & 10.29 & 10.20 & 10.03 & 10.31 & 10.36 & 10.37 & 10.13 & 10.18$\pm$0.13 \\
        \bottomrule
        \end{tabular}
    }
    \label{tab:stream_order}
\end{table}

\subsection{Alternative Strategies for Updating Target Domain Clusters
}
In the manuscript, we presented target domain clustering through pseudo labeling. A temporal consistency approach is adopted to filter out confident samples to update target clusters. In this section, we discuss two alternative strategies for updating target domain clusters. Firstly, each target cluster can be updated with all samples assigned with respective pseudo label~(Without Filtering). This strategy will introduce many noisy samples into cluster updating and potentially harm test-time feature learning. Secondly, we use a soft assignment of testing samples to each target cluster to update target clusters~(Soft Assignment). This strategy is equivalent to fitting a mixture of Gaussian through EM algorithm. Finally, we compare these two alternative strategies with our temporal consistency based filtering approach. The results are presented in Tab.~\ref{tab:target_clust_update}. We find the results with temporal consistency based filtering outperforms the other two strategies on 13 out of 15 categories of corruptions, suggesting pseudo label filtering is necessary for estimating more accurate target clusters.

\begin{table}[]
    \centering
    \caption{Comparison of alternative strategies for updating target domain clusters.  }
    \resizebox{\linewidth}{!}{
        \begin{tabular}{l|ccccccccccccccc|c}
        \toprule
            Strategy & Bird & Contr & Defoc & Elast & Fog & Frost & Gauss & Glass & Impul & Jpeg & Motn & Pixel & Shot & Snow & Zoom & Avg \\
        \midrule
            i. Without filtering & 7.19 & 8.98 & 9.29 & 17.28 & 11.90 & 11.72 & 17.19 & 22.47 & 20.83 & 12.27 & 10.11 & 12.39 & 13.85 & 11.56 & 7.97 & 13.00 \\
            ii. Soft Assignment & 6.77 & \textbf{8.02} & 7.93 & \textbf{14.77} & 10.87 & 10.68 & 13.65 & 18.69 & 17.58 & 11.26 & 9.33 & 9.54 & 11.70 & 10.56 & 6.93 & 11.22 \\
            Filtering~(Ours)  & \textbf{6.41} & 8.05 & \textbf{7.85} & 14.81 & \textbf{10.28} & \textbf{10.51} & \textbf{13.06} & \textbf{18.36} & \textbf{17.35} & \textbf{10.80} & \textbf{8.97} & \textbf{9.34} & \textbf{11.61} & \textbf{10.01} & \textbf{6.68} & \textbf{10.94} \\
        \bottomrule
        \end{tabular}
    }
    \label{tab:target_clust_update}
\end{table}

\subsection{Sensitivity to Hyperparameters}
We evaluate the sensitivity to two thresholds during pseudo label filtering, namely the temporal smoothness threshold $\tau_{TC}$ and posterior threshold $\tau_{PP}$. $\tau_{TC}$ controls how much the maximal probability deviate from the historical exponential moving average. If the current value is lower than the ema below a threshold, we believe the prediction is not confident and the sample should be excluded from estimating target domain cluster. $\tau_{PP}$ controls the the minimal maximal probability and below this threshold is considered as not confident enough. We evaluate $\tau_{TC}$in the interval between 0 and -1.0 and $\tau_{PP}$ in the interval from 0.5 to 0.95 with results on CIFAR10-C level 5 glass blur corruption presented in Tab.~\ref{tab:Thresholds}. We draw the following conclusions on the evaluations. i) There is a wide range of hyperparameters that give stable performance, e.g. $\tau_{TC}\in[0.5,0.0.9]$ and $\tau_{PP}\in[-0.0001,-0.01]$. ii) When temporal consistency filtering is turn off, i.e. $\tau_{TC}=-1.0$, because the probability is normalized to between 0 and 1, the performance drops substantially, suggesting the necessity to apply temporal consistency filtering.

\begin{table}[ht]
    \caption{Evaluation of pseudo labeling thresholds on CIFAR10-C level 5 glass blur corruption. Numbers are reported as classification error (\%).}
    \centering
    \begin{tabular}{c|ccccccc}
    \toprule
        $\tau_{TC}\backslash \tau_{PP}$ & 0.5 & 0.6 & 0.7 & 0.8 & 0.9 & 0.95 \\
        \midrule
        0.0 & 23.03 & 22.26 & 21.96 & 22.50 & 21.14 & 28.55 \\
        -0.0001 & 20.03 & 20.53 & 20.45 & 20.40 & 19.49 & 27.00 \\
        -0.001 & 19.66 & 20.51 & 19.49 & 20.48 & \textbf{19.42} & 26.83 \\
        -0.01 & 20.71 & 20.78 & 20.73 & 20.65 & 20.29 & 27.58 \\
        -0.1 & 24.10 & 21.47 & 21.46 & 22.36 & 21.45 & 28.71 \\
        -1.0 & 30.75 & 24.08 & 23.40 & 24.33 & 22.21 & 28.77 \\
    \bottomrule
    \end{tabular}
    \label{tab:Thresholds}
\end{table}

\subsection{Improvement by KL-Divergence}
Minimizing KL-Divergence between two Gaussian distributions is equivalent to matching the first two moments of the true distributions~\cite{7528140}. TFA or TTT++ aligns the first two moments through minimizing the L2/F norm, referred to as L2 alignment hereafter. Although L2 alignment is derived from Central Moment Discrepancy~\cite{ZELLINGER2019174}, the original CMD advocates a higher order moment matching and the weight applied to each moment is hard to estimate on real-world datasets. An empirical weight could be applied to balance the mean and covariance terms in TTT++, at the cost of introducing additional hyperparameters. We also provide a comparison between KL-Divergence and L2 alignment on CIFAR10-C level 5 snow corruption in Tab.~\ref{tab:KLDvsL2} using the original code released by TTT++. The performance gap empirically demonstrates the superiority of KL-Divergence. Nevertheless, we believe a theoretical analysis into why KL-Divergence is superior under test-time training would be inspirational and we leave it for future work.

\begin{table}[ht]
    \caption{Comparing KL-Divergence and L2 alignment as test-time training loss with the original code released by TTT++ (Y-M) on CIFAR10  level 5 snow corruption.}
    \centering
    \begin{tabular}{c|c}
    \toprule
    Feature Alignment Strategy & Error (\%) \\
    \midrule
        L2 alignment (original TTT++) & 9.85 \\
        KL-Divergence & 8.43 \\
    \bottomrule
    \end{tabular}
    \label{tab:KLDvsL2}
\end{table}

\section{Limitations and Failure Cases}

We discuss the limitations of our method from two perspectives. First, we point out that TTAC implements backpropagation to update models at test stage, therefore additional computation overhead is required. Specifically, as Tab.~\ref{tab:QueueTimeAnalysis}, we carried out additional evaluations on the per-sample wall clock time. Basically, we discovered that TTAC is 2-5 times computationally more expensive than BN and TENT. However, contrary to usual recognition, BN and TENT are also very expensive compared with no adaptation at all. Eventually, most test-time training methods might require an additional device for test-time adaptation. 

We further discuss the limitations on test-time training under more severe corruptions. Specifically, we evaluate TENT, SHOT and TTAC under 1-5 levels of corruptions on CIFAR10-C with results reported in Tab.~\ref{tab:CorruptLevel}. We observe generally a drop of performance from 1-5 level of corruption. Despite consistently outperforming TENT and SHOT at all levels of corruptions, TTAC's performance at higher corruption levels are relatively worse, suggesting more attention must be paid to more severely corrupted scenarios. 

\begin{table}[ht]
    \caption{Classification error under different levels of snow corruption on CIFAR10-C dataset.}
    \centering
    \begin{tabular}{c|ccccc}
    \toprule
        Level & 1 & 2 & 3 & 4 & 5 \\
    \midrule
        TEST & 9.46 & 18.34 & 16.89 & 19.31 & 21.93 \\
        TENT & 8.76 & 11.39 & 13.37 & 15.18 & 13.93 \\
        SHOT & 8.70 & 11.21 & 13.16 & 15.12 & 13.76 \\
        TTAC & 6.54 & 8.19 & 9.82 & 10.61 & 9.98 \\
    \midrule
    \end{tabular}
    \label{tab:CorruptLevel}
\end{table}

\section{Detailed results}

We further provide details of test-time training on CIFAR10-C, CIFAR100-C and ModelNet40-C datasets in Tab.~\ref{tab:cifar10c},~\ref{tab:cifar100c} and ~\ref{tab:modelnet40c} respectively. The results in Tab.~\ref{tab:cifar10c} and ~\ref{tab:cifar100c} suggest TTAC has a powerful ability to adapt to the corrupted images, and obtains the state-of-the-art performances on almost all corruption categories. 

\begin{table}[!htb]
    \centering
    \caption{The results of CIFAR10-C under the sTTT protocol}
    \resizebox{\linewidth}{!}{
        \begin{tabular}{l|ccccccccccccccc|c}
        \toprule
        Method & Bird & Contr & Defoc & Elast & Fog & Frost & Gauss & Glass & Impul & Jpeg & Motn & Pixel & Shot & Snow & Zoom & Avg \\
        \midrule
        TEST & 7.00 & 13.28 & 11.84 & 23.38 & 29.42 & 28.25 & 48.73 & 50.79 & 57.01 & 19.46 & 23.38 & 47.88 & 44.00 & 21.93 & 10.84 & 29.15 \\
        BN   & 8.21 & 8.36  & 9.73  & 19.43 & 20.16 & 13.72 & 17.46 & 26.34 & 28.11 & 14.00 & 13.90 & 12.22 & 16.64 & 16.00 & 8.03  & 15.49 \\
        TENT & 8.22 & 8.07  & 9.93  & 18.29 & 15.65 & 14.14 & 16.60 & 24.10 & 25.80 & 13.39 & 12.34 & 11.06 & 14.75 & 13.87 & 7.87  & 14.27 \\
        T3A  & 8.33 & 8.70 & 9.70 & 19.51 & 20.26 & 13.83 & 17.27 & 25.61 & 27.63 & 14.05 & 14.26 & 12.12 & 16.37 & 15.78 & 8.13 & 15.44 \\
        SHOT & 7.58 & 7.78  & 9.12  & 17.76 & 16.90 & 12.56 & 15.99 & 23.30 & 24.99 & 13.19 & 12.59 & 11.37 & 14.85 & 13.75 & 7.51  & 13.95 \\
        TTT++ & 7.70 & 7.91 & 9.24 & 17.55 & 16.39 & 12.74 & 15.49 & 22.57 & 22.86 & 13.02 & 12.52 & 11.46 & 14.45 & 13.90 & 7.51 & 13.69 \\
        TTAC (Ours) & 6.41 & 8.05 & 7.85 & 14.81 & \textbf{10.28} & \textbf{10.51} & \textbf{13.06} & 18.36 & \textbf{17.35} & \textbf{10.80} & 8.97 & 9.34 & \textbf{11.61} & \textbf{10.01} & \textbf{6.68} & \textbf{10.94} \\
        TTAC+SHOT (Ours) & \textbf{6.37} & \textbf{6.98} & \textbf{7.79} & \textbf{14.80} & 11.04 & 10.52 & 13.58 & \textbf{18.34} & 17.68 & 10.94 & \textbf{8.93} & \textbf{9.20} & 11.81 & 10.01 & 6.79 & 10.99 \\
        \bottomrule
        \end{tabular}
    }
    \label{tab:cifar10c}
\end{table}

\begin{table}[!htb]
    \centering
    \caption{The results of CIFAR100-C under the sTTT protocol}
    \resizebox{\linewidth}{!}{
        \begin{tabular}{l|ccccccccccccccc|c}
        \toprule
        Method & Bird & Contr & Defoc & Elast & Fog & Frost & Gauss & Glass & Impul & Jpeg & Motn & Pixel & Shot & Snow & Zoom & Avg \\
        \midrule
        TEST & 28.84 & 50.87 & 39.61 & 59.53 & 68.10 & 60.21 & 80.77 & 82.27 & 87.75 & 49.98 & 54.20 & 72.27 & 77.84 & 54.57 & 38.36 & 60.34 \\
        BN   & 31.78 & 33.06 & 33.86 & 48.65 & 54.23 & 42.28 & 48.02 & 57.08 & 60.14 & 39.09 & 40.72 & 37.76 & 45.83 & 46.31 & 31.91 & 43.38 \\
        TENT & 30.45 & 31.47 & 32.48 & 45.84 & 44.85 & 41.39 & 45.59 & 52.31 & 56.16 & 38.94 & 38.41 & 35.55 & 43.40 & 42.89 & 31.10 & 40.72 \\
        T3A   & 31.66 & 32.63 & 33.62 & 47.60 & 53.06 & 41.95 & 46.63 & 55.51 & 58.92 & 38.89 & 40.26 & 37.21 & 45.32 & 46.08 & 31.43 & 42.72 \\
        SHOT & 29.36 & \textbf{30.49} & 31.33 & 43.41 & 45.14 & 39.31 & 43.35 & 50.98 & 53.75 & 36.07 & 36.11 & 34.54 & 42.16 & 40.99 & 29.52 & 39.10 \\
        TTT++ & 30.79 & 31.48 & 33.04 & 44.95 & 47.74 & 40.19 & 43.94 & 52.06 & 54.08 & 37.26 & 38.10 & 35.40 & 42.28 & 42.97 & 30.58 & 40.32 \\
        TTAC (Ours) & 28.13 & 32.55 & 29.45 & 41.54 & 39.07 & 36.95 & 40.01 & 48.30  & 49.21 & 34.55 & 33.29 & 32.69 & 38.62 & 37.69 & 27.61 & 36.64 \\
        TTAC+SHOT (Ours) & \textbf{27.73} & 32.19 & \textbf{29.25} & \textbf{41.26} & \textbf{38.67} & \textbf{36.67} & \textbf{40.01} & \textbf{47.87} & \textbf{49.21} & \textbf{34.13} & \textbf{32.98} & \textbf{32.52} & \textbf{38.62} & \textbf{37.35} & \textbf{27.36} & \textbf{36.39} \\
        \bottomrule
        \end{tabular}
    }
    \label{tab:cifar100c}
\end{table}

\begin{table}[!ht]
    \centering
    \caption{The results of ModelNet40-C under the sTTT protocol}
    \resizebox{\linewidth}{!}{
        \begin{tabular}{l|ccccccccccccccc|c}
        \toprule
        Method & Background & Cutout & Density Inc. & Density Dec. & Inv. RBF & RBF & FFD & Gaussian & Impulse & LiDAR & Occlusion & Rotation & Shear & Uniform & Upsampling & Avg\\
        \midrule
        TEST & 57.41 & 23.82 & 16.17 & 27.59 & 21.19 & 22.85 & 19.89 & 27.07 & 37.48 & 85.21 & 65.24 & 41.61 & 16.33 & 22.93 & 34.44 & 34.62 \\
        BN   & 52.88 & 18.07 & 13.25 & 20.42 & 16.57 & 17.50 & 17.75 & 17.30 & 18.60 & 70.75 & 58.51 & 26.94 & 14.51 & 15.48 & 19.37 & 26.53 \\
        TENT & 51.94 & 17.38 & 13.25 & 17.99 & 14.14 & 16.65 & 15.68 & 16.49 & 17.10 & 81.44 & 64.18 & 22.33 & 13.29 & 14.59 & 19.25 & 26.38 \\
        T3A   & 52.51 & 16.37 & 13.09 & 18.23 & 14.26 & 15.48 & 15.88 & 14.14 & 15.68 & 69.12 & 54.82 & 24.80 & 13.01 & 14.14 & 17.06 & 24.57 \\
        SHOT & \textbf{15.64} & \textbf{14.34} & 12.24 & \textbf{15.48} & 13.37 & \textbf{13.82} & \textbf{12.64} & 13.13 & \textbf{13.43} & 66.05 & \textbf{47.41} & \textbf{18.80} & \textbf{11.79} & 12.44 & 15.11 & 19.71 \\
        TTAC (Ours) & 24.88 & 17.14 & 12.44 & 19.12 & 15.07 & 16.29 & 16.45 & 14.95 & 16.37 & 63.49 & 52.19 & 22.41 & 13.70 & 13.78 & 16.21 & 22.30 \\
        TTAC+SHOT (Ours) & 18.67 & 14.89 & \textbf{10.88} & 15.58 & \textbf{13.12} & 14.19 & 14.04 & \textbf{12.15} & 14.08 & \textbf{57.35} & 47.48 & 18.93 & 11.99 & \textbf{11.92} & \textbf{12.88} & \textbf{19.21} \\
        \bottomrule
        \end{tabular}
    }
    \label{tab:modelnet40c}
\end{table}

\end{document}